\pdfoutput=1

\documentclass{article} % For LaTeX2e
 
\usepackage{nips14submit_e,times}
\usepackage{hyperref}
\usepackage{url}

 \usepackage{amsmath,amssymb, amsthm}

\usepackage{multirow}
\usepackage{amsfonts}
\usepackage{layout}
\usepackage{amsmath}
\usepackage{amsthm}
\usepackage{amssymb} 
\usepackage{url}
\usepackage{color}
\usepackage{graphicx}
\usepackage{algorithmic}
\usepackage{algorithm}
\usepackage{verbatim}

 \newcommand{\xtilde}{\tilde x}

 \newcommand{\xbar}{\bar x}

\newcommand\tagthis{\addtocounter{equation}{1}\tag{\theequation}}

\newcommand{\eqdef}{\stackrel{\text{def}}{=}}

\newcommand{\R}{\mathbb{R}}

\DeclareMathOperator{\prox}{prox}         

\DeclareMathOperator{\Exp}{\mathbf{E}}           % expectation
\DeclareMathOperator{\dom}{dom}         % domain
\newtheorem{assumption}{Assumption}
\newtheorem{lemma}{Lemma}

\newtheorem{theorem}{Theorem}

\newtheorem{remark}{Remark}

\theoremstyle{plain}

\theoremstyle{definition}

\title{mS2GD: Mini-Batch  Semi-Stochastic Gradient Descent in the Proximal Setting}

\author{
Jakub Kone\v{c}n\'{y}\\
University of Edinburgh\\
United Kingdom, EH9 3FD \\
\texttt{J.Konecny@sms.ed.ac.uk} \\
\And
Jie Liu\\
Lehigh University\\
Bethlehem, PA 18015 \\
\texttt{jie.liu@lehigh.edu} \\
\And
Peter Richt\'{a}rik \\
University of Edinburgh\\
United Kingdom, EH9 3FD \\
\texttt{peter.richtarik@ed.ac.uk} 
\And
Martin Tak\'a\v{c}\\
Lehigh University\\
Bethlehem, PA 18015 \\
\texttt{takac.mt@gmail.com} \\
}

\nipsfinalcopy % Uncomment for camera-ready version

\newcommand{\stepsize}{ h}
\newcommand{\decrease}{\rho}
\newcommand{\primal}{P}

\newcommand{\z}{{z}}
\newcommand{\x}{{x}}
\newcommand{\y}{{y}}
\newcommand{\napad}[1]{\hrule {\small \it #1} \hrule }
\renewcommand{\xbar}{{\bar x}}
\renewcommand{\xtilde}{{\tilde x}}
\newcommand{\setn}{[n]}

\begin{document}

\maketitle

\begin{abstract} 
We propose a mini-batching scheme for improving the theoretical complexity and practical performance of semi-stochastic gradient descent applied to the problem of minimizing a strongly convex composite function represented as the sum of an average of a large number of smooth convex functions, and simple nonsmooth convex function. Our method first performs a deterministic step (computation of the gradient of the objective function  at the starting point), followed by a large number of stochastic steps. The process is repeated a few times with the last iterate becoming the new starting point. The novelty of our method is in introduction of mini-batching into the computation of stochastic steps. In each step, instead of choosing a single function, we sample $b$ functions, compute their gradients, and compute the direction based on this. We analyze the complexity of the method and show that the method benefits from two speedup effects. First,  we prove that as long as $b$ is below a certain threshold, we can 
reach predefined accuracy with less overall work than without mini-batching. Second, our mini-batching scheme admits a simple parallel implementation, and hence is suitable for further acceleration by parallelization.
\end{abstract}

\section{Introduction}

The problem we are interested in is to minimize a sum of two convex functions,
\begin{equation}\label{Px1}
\min_{\x\in\R^d} \{\primal(\x) := f(\x) + R(\x)\},
\end{equation}
where $f$ is the average of a large number of smooth convex  functions $f_i(\x)$,  i.e.,
\begin{equation}\label{Px2}
f(\x) = \frac{1}{n} \sum_{i=1}^n f_i(\x),
\end{equation}
We further make the following assumptions:

\begin{assumption}\label{ass1}
The regularizer $R:\R^d\to \R\cup \{+\infty\}$ is  convex and closed. The functions  $f_i:\R^d\to \R$ are differentiable and have Lipschitz continuous gradients with constant $L > 0$. That is, 
\begin{equation*}
\| \nabla f _i(\x)-\nabla f _i(\y) \|\leq L \|\x-\y\|,
\end{equation*}
for all $x,y\in \R^d$, where $\|\cdot\|$ is L2 norm.
\end{assumption}

Hence,  the gradient of $f$ is also Lipschitz continuous with the same constant $L$. %The Lipschitz continuity of the gradients also suggest the following holds for any function $f$, such as $F$ and $f_i$,
%\begin{equation}\label{lipcont}
%f(y)-f(x) \leq \nabla f(x)^T(y-x) + \frac{L_f}2\|y-x\|^2, \forall %x,y \in \dom(f)
%\end{equation}
%where $L_f$ is the Lipschitz constant for $f$.

\begin{assumption}\label{ass2}
 $\primal$ is strongly convex with parameter $\mu>0$, i.e.,  $\forall \x,\y \in \dom(R)$,
\begin{equation}\label{strongconv}
\primal(\y) \geq \primal(\x) + \xi^T(\y-\x) + \frac\mu2 \|\y-\x\|^2, \ \ \forall \xi \in \partial{\primal(\x)},
\end{equation}
where $\partial \primal(\x)$ is the subdifferential of $\primal$ at $\x$.
\end{assumption}

Let $\mu_f \geq 0$ and $\mu_R\geq 0$ be the strong convexity parameters of $f$ and $R$, respectively (we allow both of them to be equal to $0$, so this is not an additional assumption). We assume that we have lower bounds available ($\nu_f\in [0,\mu_f]$ and $\nu_R\in[0,\mu_R]$).

\paragraph{Related work}

There has been intensive interest and activity in solving problems of \eqref{Px1} in the past years. An algorithm that made its way into many applied areas is FISTA \cite{fista}. However, this method is impractical in large-scale setting (big $n$) as it needs to process all $n$ functions in each iteration. Two classes of methods address this issue -- randomized coordinate descent methods \cite{nesterovRCDM, richtarik, richtarikBigData, necoara2014random, approx,shalev2013stochastic, marecek2014distributed,necoara2013distributed, richtarik2013distributed, fercoq2014fast} and stochastic gradient methods \cite{zhangsgd, nemirovskisgd, jaggi2014communication, takac2013ICML}. This brief paper is closely related to the works on stochastic gradient methods with a technique of explicit variance reduction of stochastic approximation of the gradient. In particular, our method is a mini-batch variant of S2GD \cite{s2gd};  the proximal setting was motivated by SVRG \cite{svrg, proxsvrg}.

A typical stochastic gradient  descent (SGD) method will randomly sample $i^{th}$ function and then update the variable $\x$ using $\nabla f_i(\x)$ --- an estimate of $\nabla f(\x)$. Important limitation of SGD is that it is inherently sequential, and thus it is difficult to parallelize them. In order to enable parallelism, mini-batching---samples multiple examples per iteration---is often employed \cite{pegasos, dekel, acceleratedmb, zhangmb,takac2013ICML,shalev2013accelerated}.

\paragraph{Our Contributions.}

In this work, we combine the variance reduction ideas for stochastic gradient methods with mini-batching. In particular,  develop and analyze mS2GD (Algorithm~\ref{algorithm2}) -- a mini-batch proximal variant of S2GD \cite{s2gd}. To the best of our knowledge, this is the first {\em mini-batch} stochastic gradient method with reduced variance for problem \eqref{Px1}.  We show that the method enjoys twofold benefit compared to previous methods. Apart from admitting a parallel implementation (and hence speedup in clocktime in an HPC environment), our results show that in order attain a  specified accuracy $\epsilon$, our mini-batching scheme can get by with less gradient evaluations. This is formalized in Theorem~\ref{thm:optimalM}, which predicts more than linear speedup up to some $b$ --- the size of the mini-batches. Another advantage, compared to \cite{proxsvrg}, is that we do not need to average the $t_k$ points $\x$ in each loop, but we instead simply continue from the last one (this is the approach employed in S2GD \cite{s2gd}).

%*******************
% Section
%*******************
\section{Proximal Algorithms}

A popular {\em proximal gradient} approach to solving \eqref{Px1}  is to form a sequence $\{\y_{k}\}$ via \[\y_{k+1} = \arg \min_{\x\in \R^d} \left[U_{k}(x) \eqdef f(\y_{k}) + \nabla f (\y_{k})^T (\x-\y_{k}) + \frac{1}{2\stepsize} \|\x-\y_{k}\|^2 + R(x)\right].\] Note that $U_{k}$ in an upper bound on $\primal$  if $\stepsize>0$ is a stepsize parameter satisfying $1/\stepsize \geq L$. This procedure can be equivalently  written using the {\em proximal operator} as follows:
\begin{equation*}
\y_{k+1} = \prox_{\stepsize R} (\y_{k} - \stepsize \nabla f (\y_{k})), \qquad \prox_R(\z) \eqdef \arg\min_{\x \in\R^d} \{ \tfrac 12  \|\x-\z\|^2 + R(\x)\}.
\end{equation*} 
In large-scale setting it is more efficient to instead consider  the \emph{stochastic proximal  gradient} approach, in which the proximal operator is applied to a stochastic gradient step:
\begin{equation}\label{eqn:prox-svrg update}
y_{k+1} = \prox_{\stepsize R}(y_{k} - \stepsize v_k),
\end{equation}
where $v_k$ is a stochastic estimate of the gradient $\nabla f(y_k)$. Of particular relevance to our work are the the SVRG \cite{svrg}, S2GD \cite{s2gd} and Prox-SVRG  \cite{proxsvrg} methods where the stochastic estimate of $\nabla f(y_k)$ is of the form
\begin{equation}\label{eq:sjs8js}
v_k = \nabla f (x) + \tfrac{1}{nq_{i_k}}(\nabla f _{i_k}(\y_{k}) - \nabla f_{i_k}(x)),
\end{equation}
where  $x$ is an ``old'' reference point for which the gradient $\nabla f(x)$ was already computed in the past, and $i_k \in \{1,2,\dots,n\}$ is a random index equal to $i$ with probability $q_i>0$.  Notice that $v_k$ is an unbiased estimate of the gradient: \[\Exp_i [v_k] \overset{\eqref{eq:sjs8js}}{=} \nabla f(x) + \sum_{i=1}^n q_i \tfrac{1}{n q_i} (\nabla f_i(y_k) - \nabla f_i(x)) \overset{\eqref{Px2}}{=} \nabla f(y_k).\]
Methods such as SVRG \cite{svrg}, S2GD \cite{s2gd} and  Prox-SVRG \cite{proxsvrg}  update the points $y_k$ in an inner loop, and the reference point $x$ in an outer loop. This ensures that $v_k$ has low variance, which ultimately leads to extremely fast convergence.

%*******************
% Section
%*******************
\section{Mini-batch S2GD}
\label{headings}

We now describe the mS2GD method (Algorithm~\ref{algorithm2}).

\begin{algorithm}[H]
\caption{mS2GD}
\label{algorithm2}
\begin{algorithmic}[1]
\STATE \textbf{Input:} $m$ (max \# of stochastic steps per epoch); $\stepsize>0$ (stepsize); $x_0 \in\R^d$ (starting point);  minibatch size $b\in[n]$ 
%probability distr.\ $Q^* := \{q_1^*,\dots,q_m^*\}$ defined by \eqref{probQ*}
\FOR {$k=0,1, 2, \dots$}
    \STATE Compute and store $g_{k} \leftarrow \nabla f(x_{k}) = \tfrac{1}{n}\sum_{i} \nabla f_i(x_{k})$ 
    \STATE Initialize the inner loop: $y_{k,0} \gets x_{k}$
    \STATE Let $t_k\leftarrow  t \in \{1,2,\dots,m\}$ with probability $q_t$ given by \eqref{probQ*}
    \FOR {$t=0$ to $t_k-1$}
	    \STATE Choose mini-batch $A_{kt}\subset [n]$ of size $b$, uniformly at random 
        \STATE Compute a stoch. estimate of $\nabla f(y_{k,t})$: $v_{k,t} \gets g_k + \frac{1}{b}\sum_{i\in A_{kt}}(\nabla f_{i}(y_{k,t}) - \nabla f_{i}(x_{k})) $
        \STATE $y_{k,t+1} \gets \prox_{\stepsize R}(y_{k,t} - \stepsize v_{k,t})$
     \ENDFOR
    \STATE  Set $x_{k+1} \leftarrow y_{k,t_k}$
\ENDFOR
\noindent
\end{algorithmic}
\end{algorithm}

The main step of our method (Step 8) is given by the update~\eqref{eqn:prox-svrg update}, however with the stochastic estimate of the gradient instead formed using a {\em mini-batch} of examples $A_{kt}\subset [n]$ of size $|A_{kt}|=b$. We run the inner loop for $t_k$ iterations, where $t_k=t\in \{1,2,\dots,m\}$ with probability $q_t$ given by
\begin{equation}\label{probQ*}
q_t \eqdef \frac1\gamma\left(
    \tfrac{ 1-\stepsize\mu_f}
         {1+\stepsize \nu_R}
  \right)^{m-t}, \quad \text{ with }\quad \gamma \eqdef  \sum_{t=1}^m 
 \left(
    \tfrac{ 1-\stepsize\mu_f}
         {1+\stepsize \nu_R}
  \right)^{m-t}.
\end{equation}

\section{Complexity Result}
\label{others}
 
 In this section, we state our main  complexity result and  comment on how to optimally choose the parameters of the method.
 
\begin{theorem}\label{s2convergence}
Let Assumptions~\ref{ass1} and~\ref{ass2} be satisfied and let $\x_* \eqdef \arg\min_\x \primal(\x)$. In addition, assume that the stepsize satisfies $0<\stepsize<\min\{\frac{1 - \stepsize \mu_f}
     {1+\stepsize \nu_R}
     \frac1{4   L_{max} \alpha(b)},\frac1{L_{max}}\}$
 and that $m$ is sufficiently large so that 
\begin{equation}\label{s2rho}
\decrease \eqdef 
\frac{   \left(
    \frac{ 1-\stepsize\mu_f}
         {1+\stepsize \nu_R}
  \right)^{m } \frac1{\mu} 
  +
  \frac{4\stepsize^2  L \alpha(b) }{1+\stepsize\nu_R} 
\left(      
 \gamma
+ 
  \left(
    \frac{ 1-\stepsize\mu_f}
         {1+ \stepsize \nu_R}
  \right)^{m-1}   
\right)  
  }
  {
  \gamma 
\stepsize
\left\{
\frac{ 1}
     {1+ \stepsize \nu_R}
-
 \frac{4 \stepsize   L \alpha(b) }{1- \stepsize\mu_f} 
\right\}
  } < 1,
\end{equation}
where $\alpha(b) = \frac{n-b}{b (n-1)} $. Then  mS2GD  has linear convergence in expectation:
\begin{equation*}
\Exp (\primal(\x_k) - \primal(\x_*)) \leq \decrease^k (\primal(\x_0)-\primal(\x_*)).
\end{equation*}
\end{theorem}

\begin{remark}
If we consider the  special case  $\nu_f=0$, $\nu_R=0$ (i.e., if the algorithm does not have any nontrivial good lower bounds on $\mu_f$ and $\mu_R$),  we obtain 
\begin{equation}\label{eq:afcewfveawfcwafcw}
 \decrease = 
\frac{    1
  }
  {
  m 
\stepsize \mu
(
1
-
  4\stepsize  L \alpha(b)  
)
  }
+
\frac{      
  4\stepsize  L \alpha(b) 
\left(      
 m
+ 
  1
\right)  
  }
  {
  m 
(
1
-
  4\stepsize   L \alpha(b)  
)
  }.
\end{equation}
In the special case when  $b=1$ we get $\alpha(b)=1$, and the rate given by \eqref{eq:afcewfveawfcwafcw} exactly recovers the rate achieved by Prox-SVRG \cite{proxsvrg} (in the case when the Lipschitz constats  $\nabla f_i$ are all equal).
%However, Prox-SVRG requires the computation of an average over $t_k$ points (our method does not), which is more demanding.

\end{remark}

\subsection{Mini-batch speedup} 

In order to be able to see the speed-up we can gain from the mini-batch strategy, and due to many parameters in the complexity result (Theorem \ref{s2convergence}) we need to fix some of the parameters. For simplicity, we will use $\nu_f$ and $\nu_R$ equal to $0$, so we can analyse \eqref{eq:afcewfveawfcwafcw} instead of \eqref{s2rho}. Let us consider the case when we also fix $k$ (number of outer iterations). Once the parameter $k$ is fixed and in order to get some $\epsilon$ accuracy, we get the value of $\decrease$ which will guarantee the result.

Let us now fix target decrease in single epoch $\decrease = \decrease_*$. For any $1 \leq b \leq n$, define $(\stepsize_*^b, m_*^b)$ to be the optimal pair stepsize-size of the inner loop, such that $\decrease < \decrease_*$. This pair is optimal in the sense that $m_*^b$ is the smallest possible --- because we are interested in minimizing the computational effort, thus minimizing $m$. If we set $b = 1$, we recover the optimal choice of parameters without mini-batching. If $m_*^b \leq m_*^1 / b$, then we can reach the same accuracy with less evaluations of gradient of a function $f_i$. The following Theorem states the formula for $\stepsize_*^b$ and $m_*^b$. Equation~\eqref{eq:vfewdfwaefvawvgfeefewafa} shows that as long as the condition $\tilde \stepsize^b \leq \frac1L$ is satisfied, $m_*^b$ is decreasing at a rate faster than $1/b$. Hence, we can attain the same accuracy with less work, compared to the case when $b = 1$.

%Let us now define two quantities: Let $m_*^1$ be the smallest number of inner iterations (if optimal step size $\stepsize_*^1$ is used) such that we satisfy  \eqref{eq:afcewfveawfcwafcw} and similarly $m_*^b$ would be the smallest number of inner iterations (if optimal step size $\stepsize_*^b$ is used). Note that $m_*^b$ is identical to $m_*^1$ if we choose $b=1$. Also note that $m_*^b$ is function of $\mu, L, b, n, \decrease$ only. The Theorem \ref{thm:optimalM} states the formula for $\stepsize_*^b$ and $m_*^b$. {\color{red} Highlight that we obtain $m_*^1$ directly from $m_*^b$ by setting $b = 1$ (if it's true)==YES, it is TRUE==.}

\begin{theorem}
\label{thm:optimalM}
Fix target $\decrease = \decrease_*$, where $\decrease$ is given by \eqref{eq:afcewfveawfcwafcw} and $\decrease_* \in (0, 1)$. Then, if we consider the mini-batch size $b$ to be fixed, the choice of stepsize $\stepsize_*^b$ and size of inner loop $m_*^b$ that minimizes the work done --- the number of gradients evaluated --- while having $\decrease \leq \decrease_*$, is given by the following formulas:
%Consider Theorem \ref{s2convergence} and $\decrease$ given by 
%\eqref{eq:afcewfveawfcwafcw}.
%Assume that we fix value $\decrease \in (0,1)$.
%Then if we consider the parameters of the problem $\mu, n, b, L$ of the problem to be fixed, then we distinguish two cases, 
%depending on the value of 
\begin{equation*}
\tilde \stepsize^b := \sqrt{ \left( \frac{1+\decrease}{\decrease\mu} \right)^2 + \frac1{4\mu\alpha(b)L}} - \frac{1+\decrease}{\decrease\mu}.
\end{equation*}
If $\tilde \stepsize^b \leq \frac1L$ then $\stepsize_*^b = \tilde \stepsize^b$ and
\begin{equation}
\label{eq:vfewdfwaefvawvgfeefewafa}
 m_*^b =\frac{ 4}{\left(\sqrt{\frac{\decrease^2\mu}{\alpha(b) L} + 4(1+\decrease)^2} - 2(1+\decrease)\right)}
 =
8 \alpha(b) L \frac{ 1 +   \decrease + \sqrt{
 \frac1{4 \alpha(b) L} \mu \decrease^2 +   (1 + \decrease)^2}}{\mu \decrease^2}.
\end{equation}
Otherwise  $\stepsize_*^b = \frac1L$
and
$m_*^b = \tfrac{L/\mu + 4 \alpha(b) }
  { \decrease -4 \alpha(b)   (1+\decrease)}.$

\end{theorem}

\section{Experiments}

\begin{figure}[H]
  \begin{minipage}[t]{0.38\linewidth} 
    \centering 
	\includegraphics[height = 3.5cm]{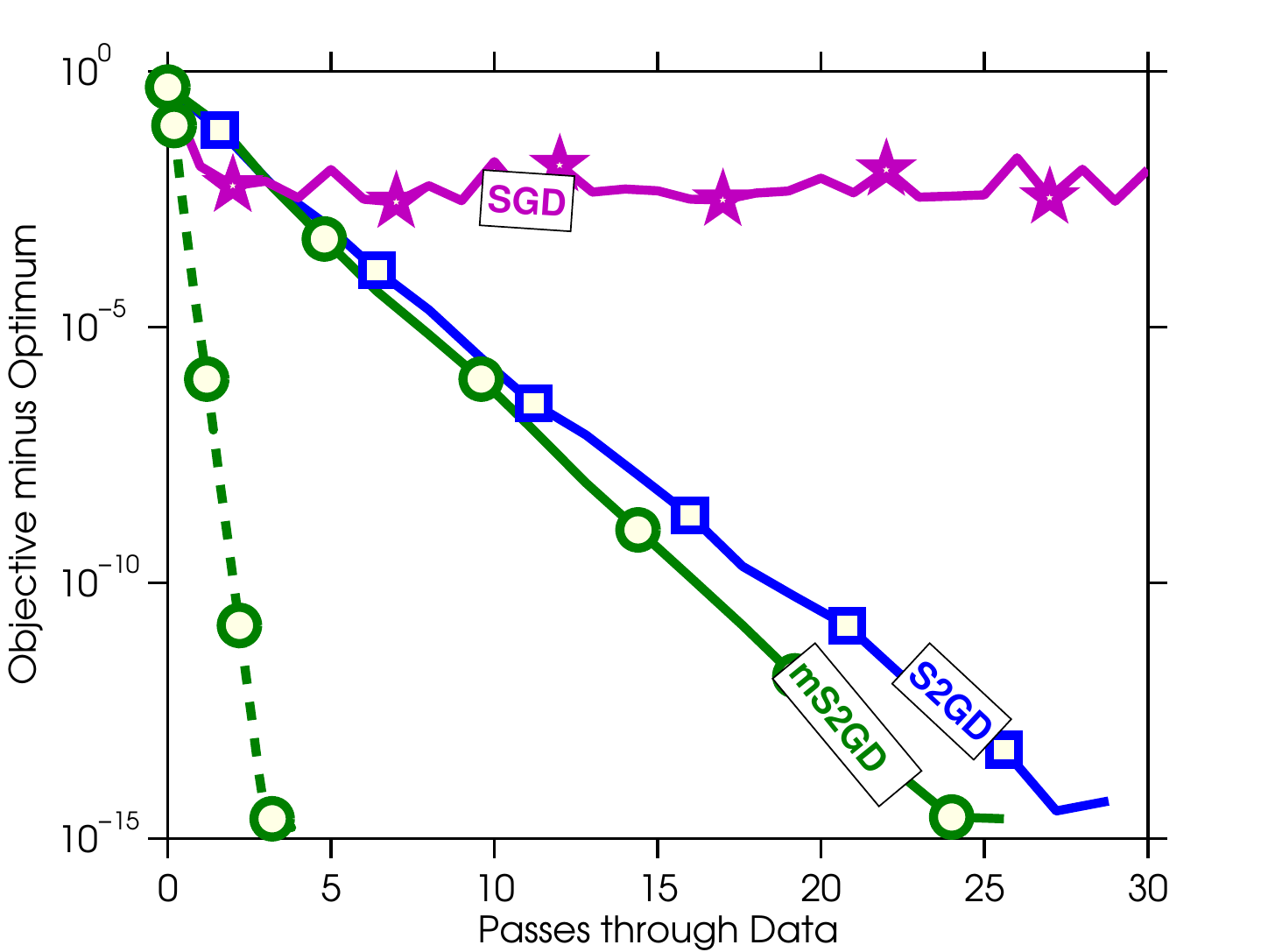}
	\caption{\footnotesize Experiment on  RCV1 dataset.}
	\label{fig:experiment}
  \end{minipage}% 
\begin{minipage}[t]{0.04\linewidth} 
$\  $
 \end{minipage}% 
  \begin{minipage}[t]{0.58\linewidth} 
    \centering 
    \includegraphics[height = 3.5cm]{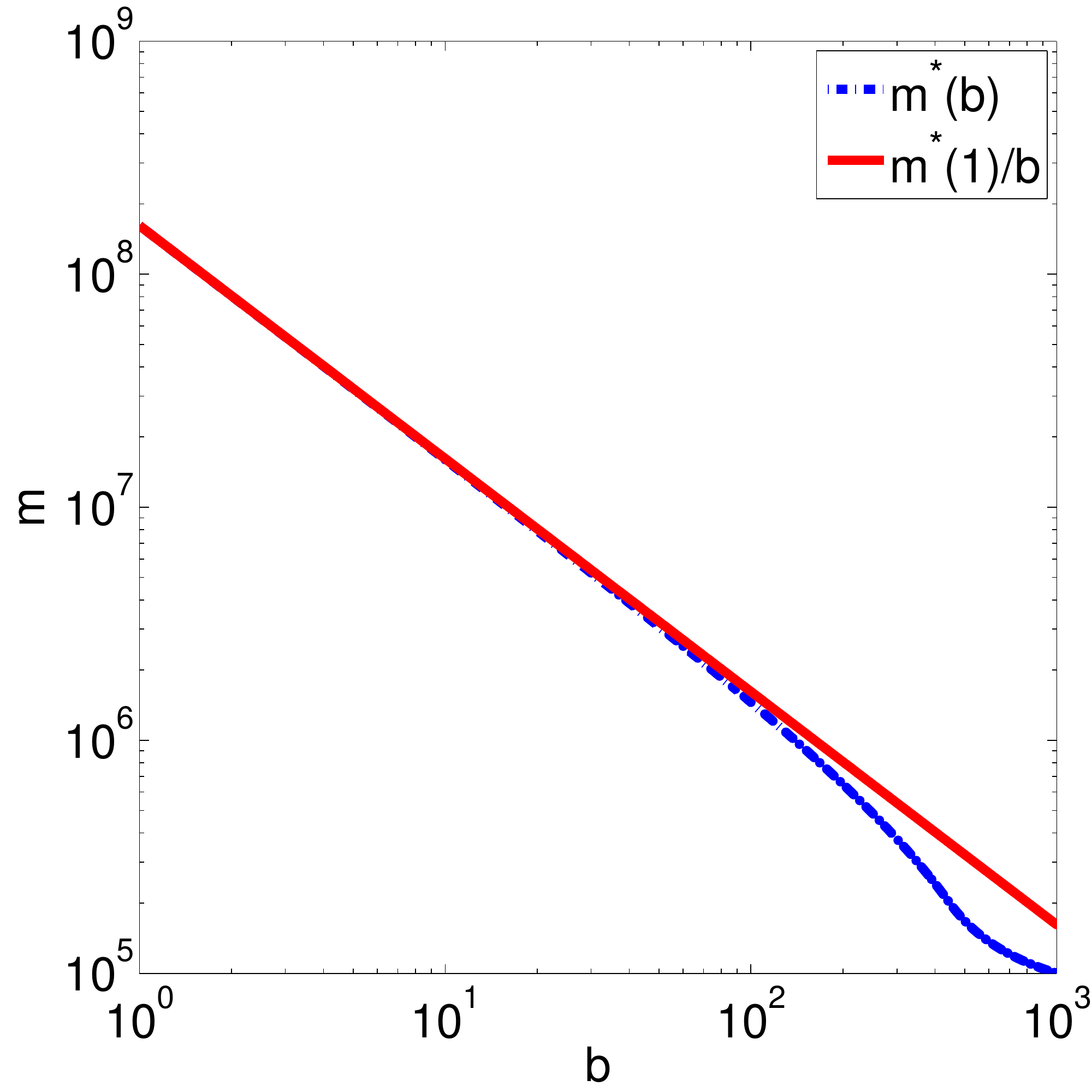}
	\includegraphics[height = 3.5cm]{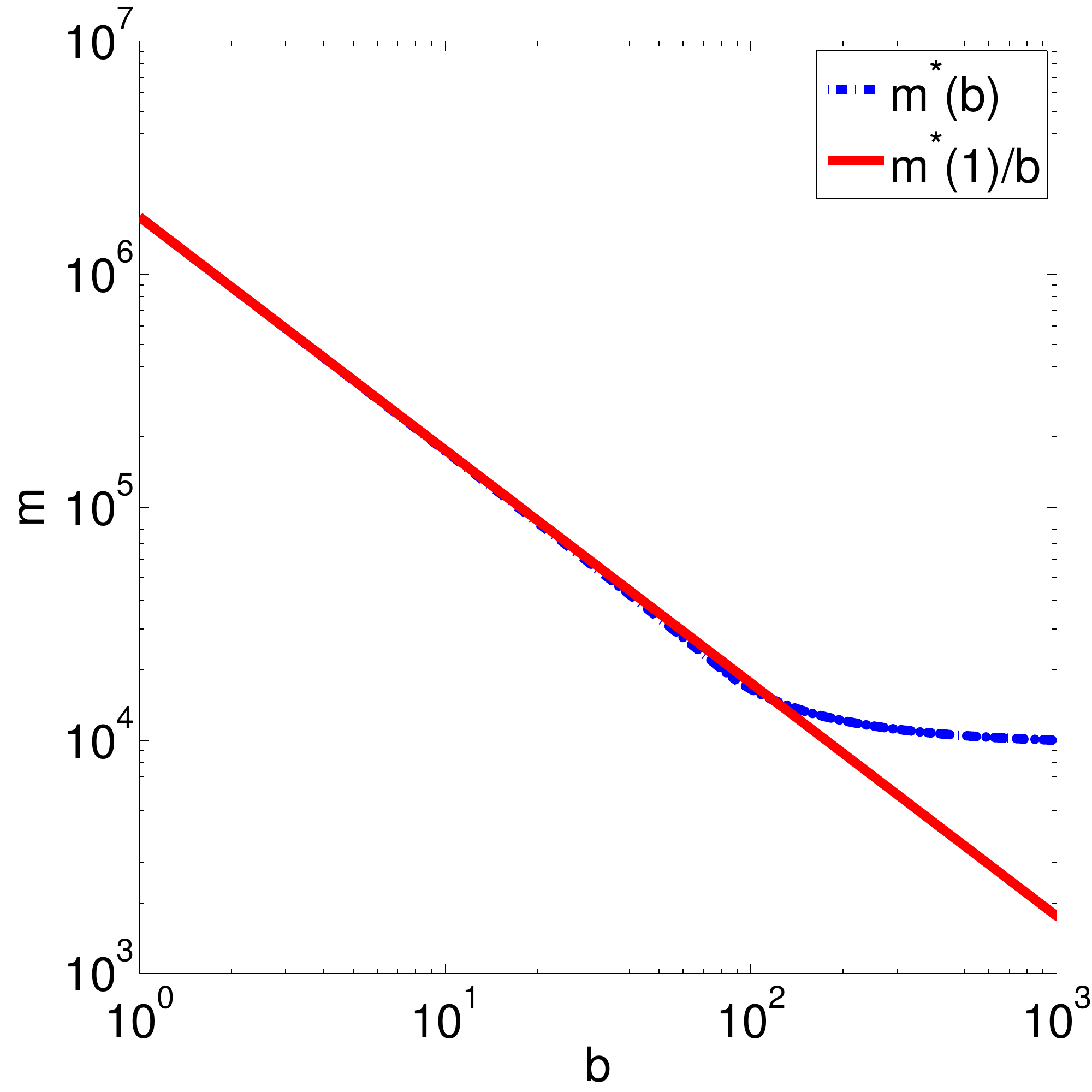}
		\\$\ $
	\caption{\footnotesize Speedup from parallelism for $\decrease=0.01$ (left) and $\decrease=0.1$ (right). Parameters: $L=1, n=1000, \mu=1/n$.}
	\label{fig:MBSpeedup}
  \end{minipage} 
\end{figure}

In this section we present a preliminary experiment, and an insight into the possible speedup by parallelism. Figure~\ref{fig:experiment} shows experiments on L2-regularized logistic regression on the RCV1 dataset.~\footnote{Available at \href{http://www.csie.ntu.edu.tw/~cjlin/libsvmtools/datasets/}{http://www.csie.ntu.edu.tw/$\sim$cjlin/libsvmtools/datasets/}.} We compare  S2GD (blue, squares) and mS2GD (green circles) with mini-batch size $b = 8$, without any parallelism. The figure demonstrates  that one can achieve the same accuracy with less work. The green dashed line is the ideal (most likely practically unachievable) result with parallelism (we divide passes through data by $b$). For comparison, we also include SGD  with constant stepsize (purple, stars), chosen in hindsight to optimize performance. Figure~\ref{fig:MBSpeedup} shows the possible speedup in terms of work done, formalized in Theorem~\ref{thm:optimalM}. Notice that up to a certain threshold, we do not need any more work to achieve the same accuracy (red straight line is ideal speedup; blue curvy line is what mS2GD achieves).

%\section{Numerical Experiments}
%
%\begin{figure}[h!] 
%\centering
%\includegraphics[width = 0.4\textwidth]{S2GDmbplot.pdf}
%\caption{Comparison of S2GD algorithm with minibatches ($b = 8$) on RCV1 dataset with S2GD and SGD. Dashed line shows ideal performance with parallelism.}
%\label{fig:experiment}
%\end{figure}
%
%The picture shows experiments on the RCV dataset \footnote{Available at \href{http://www.csie.ntu.edu.tw/~cjlin/libsvmtools/datasets/}{http://www.csie.ntu.edu.tw/$\sim$cjlin/libsvmtools/datasets/}.}. The Figure compares the S2GD algorithm, with our current algorithm (S2GDmb) with mini-batch size $b = 8$, without any parallelism. This demonstrates, that one can achieve the same accuracy with less work. The dashed line is ideal (most likely practically unachievable) result with parallelism. For comparison, we also show the SGD algorithm with constant stepsize, chosen in hindsight to optimize performance.

\bibliographystyle{plain} 
\bibliography{literature}

\end{document}